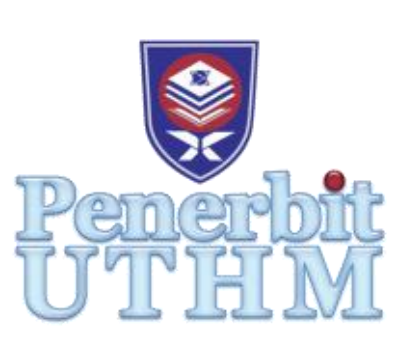

# A Survey on Multi-Resident Activity Recognition in Smart Environments

**Farhad MortezaPour Shiri[1*], Thinagaran Perumal[1], Norwati Mustapha[1], Raihani Mohamed[1], Mohd Anuaruddin Ahmadon[2], and Shingo Yamaguchi[2]**

[1]Universiti Putra Malaysia, Serdang, Selangor, MALAYSIA

[2]Yamaguchi University, Ube, Yamaguchi, JAPAN

*Corresponding Author

**Abstract**:
Human activity recognition (HAR) is a rapidly growing field that utilizes smart devices, sensors, and algorithms to automatically classify and identify the actions of individuals within a given environment. These systems have a wide range of applications, including assisting with caring tasks, increasing security, and improving energy efficiency. However, there are several challenges that must be addressed in order to effectively utilize HAR systems in multi-resident environments. One of the key challenges is accurately associating sensor observations with the identities of the individuals involved, which can be particularly difficult when residents are engaging in complex and collaborative activities. This paper provides a brief overview of the design and implementation of HAR systems, including a summary of the various data collection devices and approaches used for human activity identification. It also reviews previous research on the use of these systems in multi-resident environments and offers conclusions on the current state of the art in the field.

## 1. Introduction

Microelectronics and Internet of Things (IoT) technologies are constantly improving, and as a result, sensors are becoming more widely used in everyday situations. These intelligent items make it simple to obtain a variety of information and numerous fields can benefit from this knowledge [1]. Human activity recognition (HAR) has been a focus of research for many years, with significant contributions to the development of various technologies such as smart homes, smart health tracking, smart offices, and smart security. Using computer systems to analyze and comprehend human activity and use assistive technologies can significantly enhance daily life. [2].

HAR systems are effective in a variety of applications, including ambient assisted living and healthcare [3]. In healthcare, they can be used to monitor and manage individuals, particularly the elderly and disabled, through recognition of human activity [4]. These systems can also be utilized in sports to automate recognition of hand movements, in security systems to verify user identity through gait analysis, in self-health management, in military applications, and in human-robot interaction through gesture recognition [5].

*Corresponding author: morteza.pour@student.upm.edu.my

Despite the numerous applications and previous research and development efforts, the HAR algorithms currently face a number of challenges. One of these challenges involves recognizing simple and complex human activities [6]. Simple human activities (SHAs) are those that can be described as a single action occurring in a brief period of time, such as standing or sitting. Complex human activities (CHAs), on the other hand, often involve multiple concurrent or overlapping actions occurring over a longer time period, such as cooking or writing. It can be difficult to identify CHAs and differentiate them from SHAs. Other challenges include accurately recognizing a wide range of complex everyday activities, balancing efficiency and privacy, meeting computational requirements for portable and embedded devices, the complexity of data interpretation [7], and a major challenge related to multi-resident environments. The increasing complexity of human activities and difficulties in multi-resident environments have resulted in a data association challenge in the real world [8]. This challenge involves accurately linking each sensor event in a multi-resident area to the person who caused it [9].

The aim of this research is to provide an overview of the current status of human activity recognition (HAR) systems in multi-resident environments. Specifically, Section 2 discusses the process of designing and implementing a HAR system. In Section 3, we delve into the concepts, challenges, datasets, and literature related to recognizing activities in multi-resident settings. Finally, the paper concludes in Section 4.

## 2. HAR Systems

The design and implementation of HAR systems is surveyed briefly in this section. Figure. 1 illustrates a high-level workflow for designing an HAR system [7]. The first step for developing a HAR system is the data collection. The data for a HAR system is acquired using various devices with embedded sensors such as smart phones, and smart watches [10]. The next step in the process is to prepare the raw data for analysis through data cleaning, normalization, and the extraction of features [5]. The third step involves selecting and training a machine learning model, taking into account the number of activities and the quantity of data available [7]. This model is then trained using the cleaned data from the preprocessing stage. Finally, the model is evaluated using metrics such as accuracy, recall, and precision [11]. Further details on each of these steps are provided below.

## 2.1 Data Collection

Sensors, actuators, and smart devices such as smart phones, smart watches [10], and smart glasses [12] are used for collecting raw data in smart environments. Figure. 2 provides an overview of various data collection methods for an HAR system. In general, the data collection techniques can be classified into two categories: vision-based techniques [13] and sensor-based methods [14].

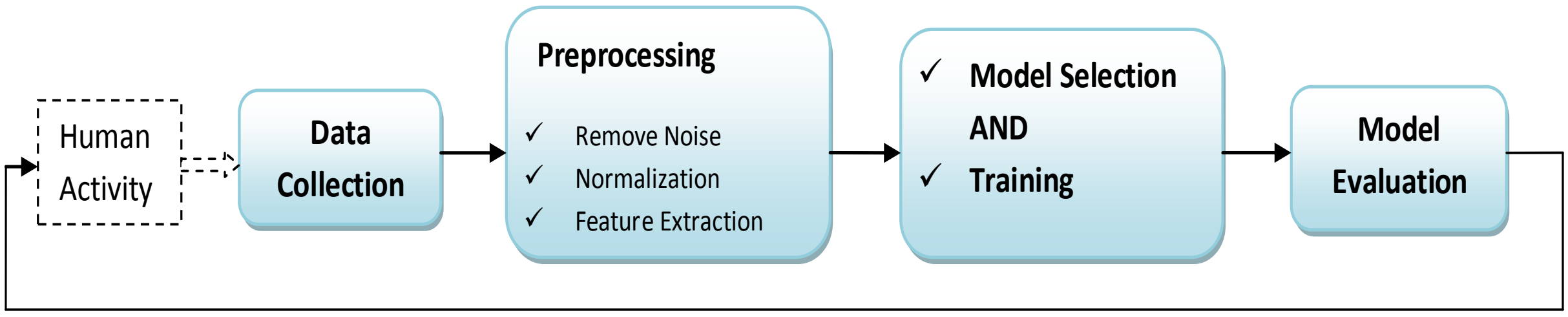

**Figure 1: HAR system workflow**

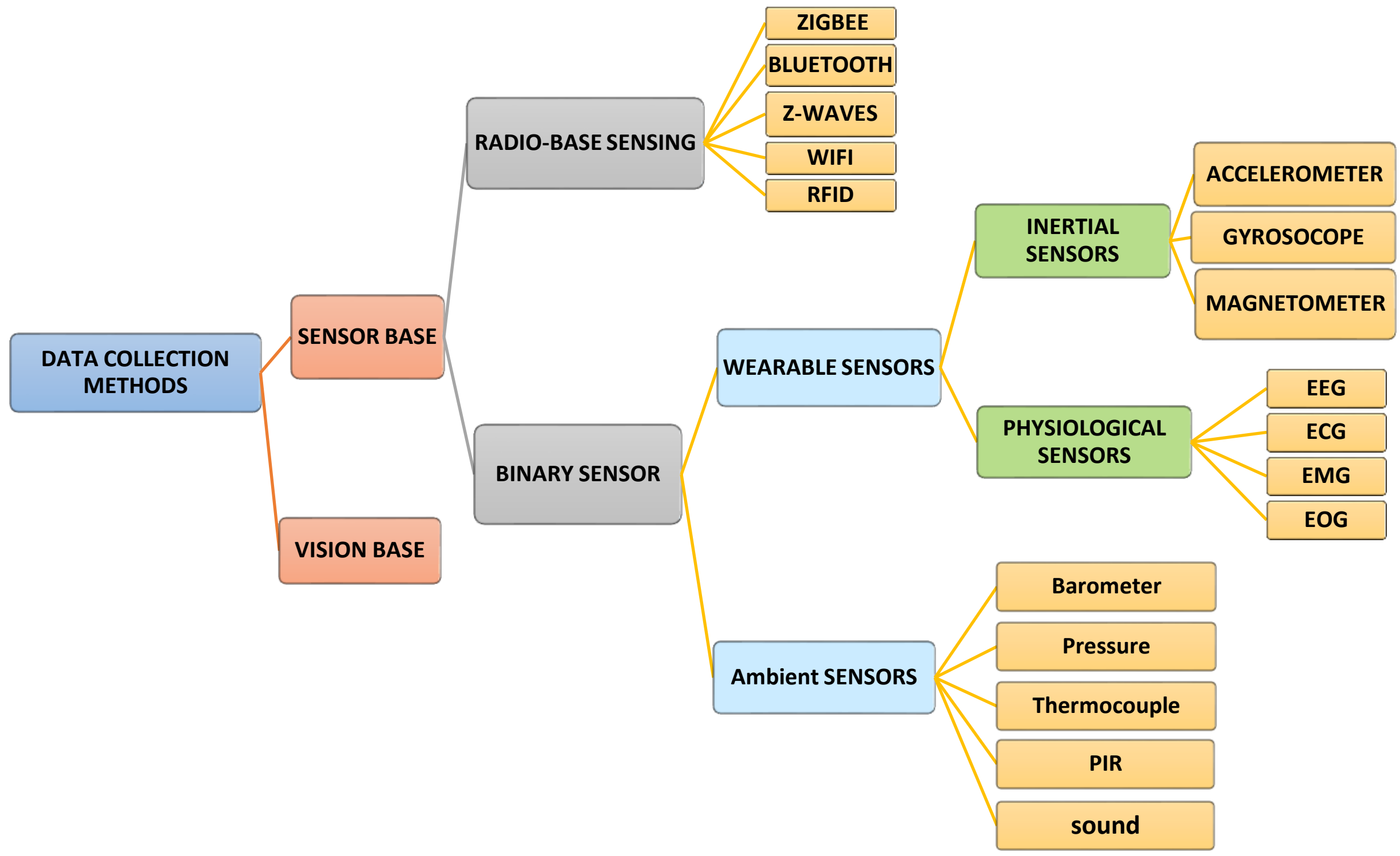

**Figure 2 : Data collection methods in HAR system**

**Vision-Based Methods**: Cameras that capture visual data are utilized by human activity recognition (HAR) systems to observe changes in the environment and monitor human behavior. A variety of cameras, including basic RGB cameras [15] and more advanced systems that utilize multiple cameras for stereo vision or employ depth cameras that measure the depth of an image using infrared light [13], are employed in vision-based techniques.

**Sensor-Based Methods**: Sensors are generally classified into two categories: radio-based sensors and binary sensors. The most common radio-based sensing systems are Bluetooth, ZigBee, Z-waves, RFID, 6LoWPAN, and WiFi [16]. Binary sensors are divided into two groups of Ambient (Environmental) sensors and wearable sensors [17]. Ambient sensors are typically installed in close proximity to collect precise information on basic environmental characteristics [7]. There are a number of common Ambient sensors such as pressure sensors, barometers, temperature sensors (Thermocouples) [17], sound sensor [18], and passive infrared (PIR) sensors [19]. One category of emerging sensor devices are wearables [3]. Wearable sensors are divided into two groups [14]: inertial sensors and physiological sensors. Inertial sensors are the most popular type of wearable technology for measuring motion and physical activities involved in daily life. Accelerometers, Gyroscopes, and Magnetic sensors are the most popular inertial sensors [20]. The last group of sensors is physiological sensors. The electrical activity of a particular bodily component is reflected in physiological signals. There are four physiological signals namely Electromyogram (EMG), Electroencephalogram (EEG), Electrocardiogram (ECG), and Electrooculogram (EOG) [21]. one of the important physiological sensor is Electroencephalogram (EEG) which is an electronic device that measures the electrical signals produced by the brain [22]. EEG sensors typically record the different electrical signals produced over time by the activity of big groups of neurons close to the surface of the brain.

## 2.2 Data Processing

Data processing is a vital step in the creation of a HAR system and involves various techniques for data cleaning, normalization of data, and extraction of features [7].

Data cleaning is the process of finding and fixing or deleting errors, anomalies, systemic problems, and inaccuracies in a dataset. Data cleaning is to guarantee that the dataset is accurate, comprehensive, and consistent so that it can be analyses successfully[23]. Data cleaning usually consist of two steps: error detection, in which difference errors are found and maybe verified by experts; and error correction, in which adjustments to the dataset are implemented (or proposed to human experts) to make the data cleaner. Real-world datasets frequently contain a variety of error types included missing values, outliers, inconsistencies, duplicates, and mislabels [24]. Missing values: A dataset may have missing value if some values are not recorded or are not known. Outliers: an outlier is observation that are significantly different from the other data points in the dataset. Duplicates: records that relate to identical real-world entities are considered duplicates. Duplicate entries in a dataset can distort analyses and produce false results. Inconsistencies: when two cells in a column have different values when they ought to have the same value, this is known as an inconsistency. For instance, in a state column, "CA" and "California," might both be present. Mislabels: When an example is labelled wrongly, it is called a mislabel.

The feature extraction process in a sensor-based HAR system generally focuses on the time and frequency domains [25]. Typically, features like as median, variance, mean, range, and skewness are extracted by time domain approaches. Features of the frequency domain technique include spectrum entropy, spectral power, peak power, and peak frequency [10]. With traditional Machine Learning techniques, features often need to be manually extracted or generated, which typically demands for specialist knowledge and a significant amount of human effort, whereas deep Learning techniques can directly extract robust features from raw data for specific purpose [26].

## 2.3 Model Selection

In order to develop a HAR system, it is necessary to select and train a classification model. This decision should take into account the quantity of activities to be recognized, the volume of data available for training, and whether computation will be performed locally or remotely [7]. Many well-known machine learning methods have been used in HAR systems for classification. These methods can be classified into wo main groups: classical machine learning (CML) methods, and deep learning (DL) techniques. Classical machine learning (CML) models have lower computational needs as well as fewer training data requirements than Deep learning (DL) models. Nevertheless, DL models are capable of identifying more complicated activities with a higher accuracy. Moreover, deep neural networks are capable of learning representative features from unprocessed data with little domain expertise [26].

### 2.3.1 Classical Machine Learning Models

Machine learning approaches employ computers to replicate human learning processes In order to acquire new information and abilities, detect current knowledge, and continually improve performance and success [27]. Machine learning approaches can generally be divided into four categories: supervised, unsupervised, semi-supervised and reinforcement [28]. Various models of machine learning have been developed in different fields. In the following, we will examine a few notable classical machine learning models used in HAR system.

#### 2.3.1.1 KNN

K Nearest Neighbor (KNN) Algorithm is a classification model which attempts to categorize the sample data point that is presented to it as a classification issue using a dataset that contains data points divided into many groups [29]. The KNN model has various merits included: It is an easy method to use since it is simple. Creating the model costs little. It is a very adaptable classification technique that works well for multimodal classes.

KNN models are used for recognizing human activity by some researchers [30, 31]. A study by [32] proposes a powerful and innovative model for human activity recognition called Reduced Kernel K-Nearest Neighbors.

#### 2.3.1.2 HMM

Hidden Markov Model (HMM) is a statistical model base on Markov process with hidden states. In HMM, the system changes over time via a series of discrete states that are not immediately observable but create observable outputs. Also, the probability of changing from one state to another is only dependent on the present state and is independent of any earlier states. A number of applications, including face identification, speech recognition, gesture recognition, bio-informatics, and gene prediction are theoretically supported by the mathematical structure of HMM [33].

Several articles are used HHM-based models for recognizing human activity in smart environments [34] [35]. A study by [9] are suggested two HMM-based models, Linked Hidden Markov Model (LHMM) and Combined-label Hidden Markov Model (CL-HMM), for recognizing human activity in smart areas with multiple residents. In [36], a Gaussian Mixture Hidden Markov Model (GMM-HMM) for human activity recognition is presented. Human activity may be described as a Markov process, and complex activity patterns can be fitted by several Gaussian density functions.

#### 2.3.1.3 Random Forest

Random Forest (RF) model is an ensemble classifier that creates several decision trees by using a randomly chosen subset of training samples and variables. The RF classifier produces trustworthy classifications by leveraging predictions obtained from a group of decision trees. Moreover, this classifier may be used to pick and rank the variables that have the best capacity to distinguish between the target classes [37].

There are various articles used models based on random forest for identifying human activities [38-40]. Principal component analysis (PCA) and random forest (RF), a hybrid technique that combines an effective approach for extracting clustering feature information with an effective algorithm for classifying data, are proposed in [41].

### 2.3.2 Deep Learning Models

The popularity of deep learning methods that use deep neural networks has grown due to the increase of high-performance computer facilities. Deep learning refers to the use of architectures with several hidden layers (deep networks) to learn various characteristics at various abstraction levels [42]. Data is passed through numerous layers by a deep learning algorithm. each layer is able of incrementally extracting features and passing the data to the next layer. Lower-level characteristics are extracted by the first layers, which are combined by later layers to create a full representation [43].

In general, Deep learning methods are divided to two models included convolutional neural networks (CNN) and Recurrent neural networks (RNN). In the following, we briefly explain about these two models and point to the articles that have used these models to recognize human activity in smart environments.

#### 2.3.2.1 CNN

Convolutional Neural Networks (CNN), is a kind of deep learning models which performs quite well and has been widely utilized in a variety of applications, including object identification, speech recognition, computer vision, video analysis, image classification, and bioinformatics[44]. The CNN identifies behaviors based on patterns, learns from the data automatically, and does not require human feature extraction. A pooling layer and several convolutional layers make up the majority of a CNN [5]. The classification operation is carried out by convolution layers, which are often followed by several fully connected layers. The goal of pooling layers is to reduce the dimensionality of input data and also to extract prominent characteristics that are invariant with regard to rotation and position[29]. Some articles are used CNN models for recognizing human activity [45]. Authors in [3] presented a HAR system that uses a Wi-Fi wearable sensor and convolutional neural networks (CNN) in order to recognize user daily activities in a smart environment. Also, in a study [46] are used Tree-Structure convolutional neural networks (TS-CNN) for multi-resident activity recognition.

#### 2.3.2.2 RNN

Recurrent Neural Networks (RNN) are a type of deep learning model, which include an internal memory. Since the output computed for the current input depends on both the input and the outcomes of previous computations, they are referred to as recurrent [29]. RNN are used for tasks that involve sequential data, such as natural language processing and speech recognition [47].

- **LSTM:** Long Short-Term Memory (LSTM) are an extension of RNN which perform significantly better than normal RNN in remembering dependencies over a lengthy period of time. LSTM contains 3 gates: input gate, forget gate, and output gate. In particular, the input gate decides how to input the current time step and how to update the system's internal state from the previous time step; the forgetting gate decides how to update the internal state from the previous time step to the internal state of the current time step; and the output gate decides how the internal state affects the system [45]. Several researchers used of LSTM method for human activity recognition [11, 48, 49]. Authors in [2] proposed a Multi-View Convolutional Neural Network and Long Short-Term Memory (CNN-LSTM) network for Radar-Based Human Activity Recognition. In [50] is used a combination of CNN-LSTM for recognizing human activity in multi-resident area.

- **GRU:** Gated Recurrent Unit (GRU) is another type of RNN and is similar to LSTM but GRU have two gates instead of three, and do not include the cell state [51]. Therefore, GRU have simpler structure than LSTM and also due to fewer tensor operations, GRU train more quickly than LSTM. However, this does not imply that it is better than LSTM. Depending on the use situation, one option may be superior to another [29]. A number of HAR researchers have demonstrated use of GRU method [5, 51]. Dong and el. [52] proposed a transformer with a bidirectional gated recurrent unit (GRU) (TRANS-BiGRU) to efficiently learn and recognize different types of activities in multiple residents environments. In [53] a multi-level neural network structure model for complex human activity recognition is proposed. It is based on the combination of Inception Neural Network and Gated Recurrent Unit (GRU).

## 2.4 Model Evaluation

The last step of developing a HAR system is evaluation of trained model. After training the model using training data, the model must be evaluated via testing data. There are several assessment measures, such as precision, accuracy, recall, and F1-measure for evaluating the trained machine learning model performance [11].

$$Accuracy = \frac{Tp + Tn}{Tp + Tn + Fp + Fn} \qquad Eq.1$$

$$Precision = \frac{Tp}{Tp + Fp} \qquad Eq.2$$

$$Recall = \frac{Tp}{Tp + Fn} \qquad Eq.3$$

$$F1 - Score = 2 \times \frac{Recall \times Precision}{Recall + Precision} \qquad Eq.4$$

Where $Tp$ = True Positive, $Tn$ = True Negative, $Fp$ = False Positive, and $Fn$ = False Negative.

## 3. Multi-Resident Activity Recognition

HAR systems have shown great success dealing with a single resident within a smart area, but real-world environment does not always have a single resident. Because of this, activity recognition for multiple residents in the smart environments should be supported multiple resident activities. Since, sensor states may not always accurately reflect a specific person's activity, recognition is a more difficult operation [50]. Also, the multi resident scenario faces other challenges such as complexity of parallel and collaborative activities [52], privacy concerns [10], need for effective integration and management of various technologies and systems involved in the recognition process such as sensors, cameras, and machine learning algorithms, and the limitations of the sensors used for activity recognition in terms of range, accuracy, and reliability [52].

In the following, we first mention the types of activities that may happen in a multi-resident environment before the stated the problems and challenges of multi resident recognition. Then, the datasets which have been published for using in this field are mentioned, and in the last part, we will examine the methods and solutions that are proposed in different literature to recognition activities of multiple residents.

## 3.1 Multi-Resident Activities

Residents of multi-resident environments not only gets involved with individual activities such as sequential, interleaving, and concurrent but also engage in parallel and collaborative activities[54].

- **Sequential Activities:** Activities that are carried out by one individual in a sequential order, one after the other, without being interwoven (e.g., make a phone call, washing hands, and then cooking) [54].

- **Interleaving Activities:** When one person switches between several different activities at once, this is called interleaving. For instance, If a resident receives a call from a friend while they are cooking, they pause their work for a bit to answer the call and then return to the kitchen to finish cooking [55].

- **Concurrent Activities:** These activities are those that a single person engages in while performing multiple tasks at once (e.g., watch TV while using phone) [55].

- **Parallel Activities:** involve multiple individuals completing tasks in the same location at the same time, such as one person making tea in the kitchen while another talks on the phone in another room [52].

- **Collaborative Activities:** involve a group working together to accomplish a common goal, either through jointly performing a task (e.g. carrying a sofa together) or working independently towards the same objective (e.g. preparing dinner together in the kitchen) [52].

### 3.2 Challenges

Monitoring multiple residents within the same environment poses serious challenges to the current state of the art HAR systems [56]. The first challenge involves recognizing simple and complex human activities. Complex human activities (CHA), often involve multiple concurrent or overlapping actions occurring over a longer period, such as cooking or writing. It can be difficult to identify complex activities and differentiate them from simple human activities (SHA). There are three major categories of CHA recognition in the literature [6]. The first approach ignores the distinctions between CHAs and SHAs and recognizes CHAs using SHA recognition techniques. Since CHAs are far more complex than SHAs, the characteristics that can be derived from SHAs are not comparable to CHAs. The second method combines SHAs to represent each CHA, where the SHAs are predetermined and manually labelled. Nevertheless, because CHAs contain many non-semantic and unlabelled components, this strategy significantly depends on domain expertise and is limited by the established SHAs. In the last scenario, CHAs are represented by latent semantics found in sensor data and by using topic models, the latent semantics are discovered. However, topic models only consider distribution data and disregard sequential data, which might help in CHA recognition.

Also, residents of multi-resident environments, despite of individual activity, can engage in parallel and Collaborative activity due to social interaction of multi-resident [52].

Another major challenge is data association, which is the process of appropriately connecting each environmental sensor event (such the opening of a refrigerator door) with the person who caused it [57]. This can be very challenging to the HAR system, because in contrast to the single-resident situation, where the sensor states directly represent the activities of a particular person, in multi-resident cases the true source of sensor readings and observations is unknown. As a result, it becomes challenging to associate each observed action to the person who is instigating the action [56]. For instance, even when two motion sensors are activated simultaneously in two distinct rooms, indicating the presence of two individuals in the environment, the HAR system will have difficulty for accurately determining which individual is in which room. Consequently, the key challenge in resident activity recognition is to associate the identities of residents to sensor observations [57].

There have been various approaches proposed in the literature for solving data association problem to multi-resident activity recognition, which can be grouped into three categories: wearable-based data association, single-model, and data-driven data association [57]. Single-model approaches rely on raw ambient sensor data to link activities to individuals, leading to implicit learning of data association during the training phase. For example, in [9] The problem of multi-resident activity recognition is investigated using two Hidden Markov Model (HMM) models, Combined label HMM (CL-HMM) and Linked HMM (LHMM). On the other hand, data-driven approaches view data association as a separate learning problem that occurs before activity classification. In [58] Authors describe an algorithm that follows each resident's position while also estimating how many people are living in a smart home without using ground-truth labeled sensor data or any other extra information. Wearable-based data association approaches require residents to wear additional sensors, such as wristbands or smartphones, to accurately associate environmental events. The authors in [59] proposed a new device to identify people in multi-resident areas, while [56] presented a multi-resident activity detection sensor network that combines Bluetooth Low Energy (BLE) signals from tags worn by occupants and passive infrared (PIR) motion sensors in the home to locate and track residents' activities.

Estimating the number of distinct residents in the same environment and track them is another challenging problem [60]. Any multi-inhabitant context-aware smart environment system must count the users present at any one time and monitor each user's movement trajectory independently and precisely as one of its primary starting stages [61].

### 3.3 Multi-Resident Dataset

There are several multi-resident activity recognition (HAR) datasets available for use. Table 1 lists a few examples.

- **ARAS**: Activity Recognition with Ambient Sensing (ARAS) dataset which includes data streams collected from two houses with multiple residents for a period of two months [62]. These houses were equipped with 20 binary sensors, such as contact sensors, pressure mats, temperature sensors, and proximity sensors, to track the residents' activities. The raw data in the ARAS dataset includes information on 27 different activities performed by the residents.
- **CASAS**: There are several multi-resident smart home datasets created by the Center of Advanced Studies in Adaptive Systems (CASAS) at Washington State University which two datasets are newer [58]. One of the datasets named TM0041 contains data recorded in a two-bedroom apartment with two old adult residents that monitored by 25 ambient sensors distributed among 8 parts of house. Another dataset is named Kyoto, a two-story house. For collecting data are used 91 sensors which are installed in bedroom, bathroom, kitchen, dining room, living room, and along hallways. Also, Magnetic door sensors are positioned on cabinet, closet, refrigerator, and front and back outdoor doors.
- **UJM**: A multi-resident dataset called SaMO- UJA, which was collected in the UJAmI Smart Lab of the University of Jaén (Spain) [63]. Data were collected from various sensors technologies and information sources included 30 binary sensors in certain objects and the 15 Bluetooth Low Energy (BLE) beacons in the space, acceleration of the resident with the wearable device and smart floor with 40 models for location. The datasets include 25 distinct types of activities.

**Table 1: Multi-resident datasets**

| **Dataset** | Name of Houses | # of Sensors | # of Activities | Duration |
|---|---|---|---|---|
| ARAS[62] | A | 20 of 7 different types | 27 | 30 days |
| ARAS[62] | B | 20 of 6 different types | 27 | 30 days |
| CASAS [58] | TM0041 | 25 | 11 | 30 days |
| CASAS [58] | Kyoto | 91 | 11 | 24 days |
| UJA [63] | SaMO- UJA | 85 | 25 | 9 days |

| SDHAR-HOME[64] | - | Wearable + Ambient sensors (35) + Positioning (7) | 18 | 60 days |
|---|---|---|---|---|
| MARBLE[65] | - | Wearable + Ambient sensors | 13 | - |

- **SDHAR-HOME:** A home where two residents and a pet live has provided as the location for the database collecting. They have also some visitors from their friends and relative. This smart home featured a collection of non-intrusive sensors to record events that happened inside home, a positioning system through triangulation using beacons, and a system to track user status using activity wristbands. The daily habits, which were measured continuously for two months and were divided into 18 distinct activity categories [64].
- **MARBLE:** The MARBLE dataset contains information from both wearable technology and ambient sensors. Smartwatches were chosen as wearables devices because of their minimal obtrusiveness, widespread use, and ability to record hand motions that may be used to identify ADLs (e.g., washing dishes). Among environmental sensors, mat (pressure) sensors can detect when people are sitting on seats or sofas, magnetic sensors can detect when drawers and doors are opened and plug sensors may detect when household appliances are being used. additionally, in order to provide indoor location, Wi-Fi access points and BLE beacons were installed [65].

## 3.4 Literature Review

Over the past 20 years, various methods for identifying human actions using sensor data have been developed. While most of these approaches are designed to work with only one person in the environment, some techniques have been created to address the more difficult task of identifying actions performed by multiple people in the same space [66]. Table 2 provides a comparison of the literature reviewed on multi-resident activity recognition.

The constraints and correlations mining engine (CACE) framework, created by the authors in reference [60], significantly improves the accuracy of recognizing complex daily activities in smart homes with multiple residents. To begin, the model generates an adjacency matrix of sensor nodes by representing the set of deployed nodes as an undirected graph. Then, the sequential importance resampling (SIR) particle filter technique is utilized to track each user's movement trajectories. After this, the state space to be explored is minimized using spatiotemporal rules that consider both deterministic correlations and statistical constraints. Finally, the Hierarchical Dynamic Bayesian Network (HDBN)-based model, which considers both micro and macro contexts, is utilized to further increase the precision of context estimation. The authors found that this approach achieved an average accuracy of 94.5 on the CASAS dataset and 95.1 on the CASE dataset (Collected Data).

In [52] to effectively learn and detect various activity types carried out by multiple residents, authors offer TRANS-BiGRU, which is a deep learning (DL) technique based on using a transformer with a Bidirectional Gated Recurrent Unit (BiGRU). The Recurrent Neural Network (RNN)-related algorithms such as Gated Recurrent Unit (GRU) and Bidirectional Gated Recurrent Unit (BiGRU) can only be computed sequentially in one of two directions: from right to left or from left to right. This technique has two drawbacks. First, the computation of time slice t depends on the outcomes of the computation at time t-1, which reduces the model's capacity for parallelism. Additionally, although the Long Short-Term Memory (LSTM)'s structure helps to some extent with the issue of long-term dependence, it is still ineffective for certain long-term dependent occurrences. A proposed transformer in [60] addresses the two mentioned problems. Provided experimental results show that proposed TRANS-BiGRU significantly outperforms other approaches with average accuracy of 89.48 on ARAS House A dataset, 90.59 on ARAS House B, and 92.86 on CASAS dataset.

**Table 2: Comparison of the reviewed literature**

| ref | Objective | Methods | Dataset | Accuracy | Limitation |
|---|---|---|---|---|---|
| **[60]** | Developing a framework to track each person's mobility trajectories. Improves the recognition accuracy of complex daily activities. | 1. Sequential Importance Resampling (SIR) particle filter technique 2. Hierarchical Dynamic Bayesian Network (HDBN) | 1.CASAS<br>2.CASE (real-world data) | 94.5%<br>95.1% | For more than two persons, the accuracy is poor. |
| **[52]** | suggested a technique for efficiently learning and identifying the various actions carried out by several residents. | A transformer with a bidirectional gated recurrent unit (TRANS-BiGRU) | ARAS (A)<br>ARAS (B)<br>CASAS | 89.48%<br>90.59%<br>92.86% | The lack of solving resident tracking problem |
| **[67]** | Extracted context information of spatial and temporal information to enhance the performance of multiple residents' activity recognition | 1. Multi-label classification namely the label combination (LC) 2. Expectation-Maximization (EM) clustering | CASAS | 98.60% | It was only done on a home with 2 residents |
| **[66]** | A new method to identify multi-inhabitant activities without labelled datasets. | Unsupervised HMM-based method | CASAS | 72.31% | insufficient support for managing collaborative activities |
| **[68]** | Improve the effectiveness of human activity recognition. Realize synchronous identification of each human and the behaviors. | 1. Split-fusion Module 2. Multi-layer perceptron (MLPs) | CASAS | 85.08% | The proposed framework is so sophisticated |
| **[69]** | Framework to model the dynamics of interaction between sensor-based environments and multiple residents. Solving identification annotation problem. | 1. Graph and Rule-Based Algorithm (GR/ED) 2. nearest neighbor standard filter (NNSF) 3. Dijkstra Algorithms | combined five single resident datasets. | 86.71% | The lack of investigate mobility models and resident tracking |

In a study by authors [67], spatial and temporal data was collected in order to recognize the activities of multiple residents in a smart home setting. They used Expectation-Maximization (EM) clustering to develop an adaptive profiling model based on potential interactions. The model was trained using multi-label classification, specifically the label combination (LC) approach, with the help of various classifiers such as random forest, K-Nearest Neighbors (KNN), Support Vector Machines (SVM), and Hidden Markov model (HMM). Out of these, the Random Forest-Label Combination (LC-RF) method yielded the highest performance with a score of 98.6 on the CASAS dataset.

A study presented in [66] presents a new approach for identifying multi resident activities without the requirement of labeled data. This is significant because the process of labeling activities can be costly and may infringe on individuals' privacy. The proposed method combines unsupervised techniques utilizing Hidden Markov Models (HMM) and ontological reasoning. Testing on the CASAS dataset demonstrated an average accuracy of 0.7231, which is similar to the performance of supervised techniques.

In [68] an end-to-end Transformer-based model named Fusion Transformer for Multi-Resident Activity Recognition (FTMAR) is proposed. The proposed approach first generates data segments with temporal information from the filtered raw data. Then uses the split-fusion for feature engineering. Two fully connected layers ultimately classify and label the output vectors of this module. The proposed approach can improve recognition performance for both residents and activities to an accuracy of 0.8855 and 0.8508 respectively, according to experimental results on the CASAS dataset.

The authors of [69] developed MoSen, a framework that models the interactions between sensor-based environments and multiple residents. One of the main challenges in using sensors to recognize activities in multi-resident environments is the identification annotation problem, which involves labeling the identities of time-series sensor events according to the residents who generated them. MoSen addresses this issue by utilizing the Graph and Rule-Based Algorithm (GR/ED). In order to create artificial multi-person datasets, the authors combined five single-occupancy datasets. Additionally, MoSen offers guidance on sensor selection and design metrics for sensor layouts in real-world deployments.

In a study [70], the authors utilize a combination of change point detection (CPD) and fuzzy c-means (FCM) for sensor event segmentation. Specifically, sensor events are classified based on their locations using the FCM method, and the CPD technique is subsequently applied to examine the transition actions to determine the segmentation sequence.

A study by [71] suggests using a Multilabel Markov Logic Network (MLN) to classify individuals based on their activity patterns and preferences. This classification considers factors such as the preferred order, length, and frequency of activities, as well as the preferred location of entities or events. The results of the experiments show that combining data-driven and knowledge-driven approaches for analyzing activities is effective.

The Attention based Deep Ensemble Learning for Activity (ADELA) system, proposed in [72], utilizes attention-based ensemble learning in a collaborative smart environment to identify the significance of events from sensor data streams based on their impact on activity recognition. The base models are assigned weights through a weighted majority calculation during the training phase and the information is stored in a trained model storage for future use in inference.

## 4. Conclusion

Human Activity Recognition (HAR) systems can greatly improve daily living by using computer systems to analyze and understand human behavior and implement assistive technologies. However, real-life environments commonly involve more than one person, with complex relationships between the residents and any visitors or family. As a result, multi-resident activity recognition is a key research issue to the development of various technologies such as smart homes, smart health tracking, smart offices, and smart security. In this paper a survey of the recent advances in HAR systems in smart environments, particularly in areas with multi residents was presented. The most important challenge in multi-residential environments is linking the actions to the people who perform them, and the complexity of parallel and collaborative activities. To overcome these challenges and achieve high accuracy, various methods have been proposed in the literature in recent years. This paper provided a brief survey of the notable works in this area.

## References

[1] Q. Li, R. Gravina, Y. Li, S. H. Alsamhi, F. Sun, and G. Fortino, "Multi-user activity recognition: Challenges and opportunities," *Information Fusion,* vol. 63, pp. 121-135, 2020, doi: 10.1016/j.inffus.2020.06.004.

[2] H. Khalid, A. Gorji, A. Bourdoux, S. Pollin, and H. Sahli, "Multi-view CNN-LSTM architecture for radar-based human activity recognition," *IEEE Access,* vol. 10, pp. 24509-24519, 2022.

[3] V. Bianchi, M. Bassoli, G. Lombardo, P. Fornacciari, M. Mordonini, and I. De Munari, "IoT wearable sensor and deep learning: An integrated approach for personalized human activity recognition in a smart home environment," *IEEE Internet of Things Journal,* vol. 6, no. 5, pp. 8553-8562, 2019.

[4] U. R. Alo, H. F. Nweke, Y. W. Teh, and G. Murtaza, "Smartphone motion sensor-based complex human activity identification using deep stacked autoencoder algorithm for enhanced smart healthcare system," *Sensors,* vol. 20, no. 21, p. 6300, 2020.

[5] S. Mekruksavanich and A. Jitpattanakul, "Deep convolutional neural network with rnns for complex activity recognition using wrist-worn wearable sensor data," *Electronics,* vol. 10, no. 14, p. 1685, 2021.

[6] L. Chen, X. Liu, L. Peng, and M. Wu, "Deep learning based multimodal complex human activity recognition using wearable devices," *Applied Intelligence,* vol. 51, no. 6, pp. 4029-4042, 2021.

[7] F. Demrozi, G. Pravadelli, A. Bihorac, and P. Rashidi, "Human activity recognition using inertial, physiological and environmental sensors: A comprehensive survey," *IEEE Access,* vol. 8, pp. 210816-210836, 2020.

[8] M. Jethanandani, T. Perumal, J.-R. Chang, A. Sharma, and Y. Bao, "Multi-Resident Activity Recognition using Multi-Label Classification in Ambient Sensing Smart Homes," in *2019 IEEE International Conference on Consumer Electronics-Taiwan (ICCE-TW)*, 2019: IEEE, pp. 1-2.

[9] A. Benmansour, A. Bouchachia, and M. Feham, "Modeling interaction in multi-resident activities," *Neurocomputing,* vol. 230, pp. 133-142, 2017.

[10] G. M. Weiss, K. Yoneda, and T. Hayajneh, "Smartphone and smartwatch-based biometrics using activities of daily living," *IEEE Access,* vol. 7, pp. 133190-133202, 2019.

[11] S. Mekruksavanich, P. Jantawong, N. Hnoohom, and A. Jitpattanakul, "Deep Learning Models for Daily Living Activity Recognition based on Wearable Inertial Sensors," in *2022 19th International Joint Conference on Computer Science and Software Engineering (JCSSE)*, 2022: IEEE, pp. 1-5.

[12] P.-E. Novac, A. Pegatoquet, B. Miramond, and C. Caquineau, "UCA-EHAR: A Dataset for Human Activity Recognition with Embedded AI on Smart Glasses," *Applied Sciences,* vol. 12, no. 8, p. 3849, 2022.

[13] D. Bouchabou, S. M. Nguyen, C. Lohr, B. LeDuc, and I. Kanellos, "A survey of human activity recognition in smart homes based on IoT sensors algorithms: Taxonomies, challenges, and opportunities with deep learning," *Sensors,* vol. 21, no. 18, p. 6037, 2021.

[14] S. Qiu *et al.*, "Multi-sensor information fusion based on machine learning for real applications in human activity recognition: State-of-the-art and research challenges," *Information Fusion,* vol. 80, pp. 241-265, 2022.

[15] N. Zerrouki, F. Harrou, Y. Sun, and A. Houacine, "Vision-based human action classification using adaptive boosting algorithm," *IEEE Sensors Journal,* vol. 18, no. 12, pp. 5115-5121, 2018.

[16] L. Babangida, T. Perumal, N. Mustapha, and R. Yaakob, "Internet of Things (IoT) Based Activity Recognition Strategies in Smart Homes: A Review," *IEEE Sensors Journal,* 2022.

[17] L. M. Dang, K. Min, H. Wang, M. J. Piran, C. H. Lee, and H. Moon, "Sensor-based and vision-based human activity recognition: A comprehensive survey," *Pattern Recognition,* vol. 108, p. 107561, 2020.

[18] E. L. Salomons, P. J. Havinga, and H. Van Leeuwen, "Inferring human activity recognition with ambient sound on wireless sensor nodes," *Sensors,* vol. 16, no. 10, p. 1586, 2016.

[19] Y. P. Raykov, E. Ozer, G. Dasika, A. Boukouvalas, and M. A. Little, "Predicting room occupancy with a single passive infrared (PIR) sensor through behavior extraction," in *Proceedings of the 2016 ACM international joint conference on pervasive and ubiquitous computing*, 2016, pp. 1016-1027.

[20] T. Tamura, "Wearable inertial sensors and their applications," in *Wearable Sensors*: Elsevier, 2014, pp. 85-104.

[21] O. Faust, Y. Hagiwara, T. J. Hong, O. S. Lih, and U. R. Acharya, "Deep learning for healthcare applications based on physiological signals: A review," *Computer methods and programs in biomedicine,* vol. 161, pp. 1-13, 2018.

[22] M. Soufineyestani, D. Dowling, and A. Khan, "Electroencephalography (EEG) technology applications and available devices," *Applied Sciences,* vol. 10, no. 21, p. 7453, 2020.

[23] J. Brownlee, *Data preparation for machine learning: data cleaning, feature selection, and data transforms in Python*. Machine Learning Mastery, 2020.

[24] P. Li, X. Rao, J. Blase, Y. Zhang, X. Chu, and C. Zhang, "CleanML: A study for evaluating the impact of data cleaning on ml classification tasks," in *2021 IEEE 37th International Conference on Data Engineering (ICDE)*, 2021: IEEE, pp. 13-24.

[25] E. Zdravevski *et al.*, "Improving activity recognition accuracy in ambient-assisted living systems by automated feature engineering," *Ieee Access,* vol. 5, pp. 5262-5280, 2017.

[26] S. Zhang *et al.*, "Deep Learning in Human Activity Recognition with Wearable Sensors: A Review on Advances," *Sensors (Basel),* vol. 22, no. 4, Feb 14 2022, doi: 10.3390/s22041476.

[27] H. Wang, C. Ma, and L. Zhou, "A brief review of machine learning and its application," in *2009 international conference on information engineering and computer science*, 2009: IEEE, pp. 1-4.

[28] I. H. Sarker, "Machine learning: Algorithms, real-world applications and research directions," *SN computer science,* vol. 2, no. 3, p. 160, 2021.

[29] S. Abbaspour, F. Fotouhi, A. Sedaghatbaf, H. Fotouhi, M. Vahabi, and M. Linden, "A Comparative Analysis of Hybrid Deep Learning Models for Human Activity Recognition," *Sensors,* vol. 20, no. 19, 2020, doi: 10.3390/s20195707.

[30] S. Mohsen, A. Elkaseer, and S. G. Scholz, "Human activity recognition using K-nearest neighbor machine learning algorithm," in *Sustainable Design and Manufacturing: Proceedings of the 8th International Conference on Sustainable Design and Manufacturing (KES-SDM 2021)*, 2022: Springer, pp. 304-313.

[31] A. Wijekoon, N. Wiratunga, S. Sani, S. Massie, and K. Cooper, "Improving kNN for human activity recognition with privileged learning using translation models," in *Case-Based Reasoning Research and Development: 26th International Conference, ICCBR 2018, Stockholm, Sweden, July 9-12, 2018, Proceedings*, 2018: Springer, pp. 448-463.

[32] Z. Liu, S. Li, J. Hao, J. Hu, and M. Pan, "An efficient and fast model reduced kernel KNN for human activity recognition," *Journal of Advanced Transportation,* vol. 2021, 2021.

[33] B. Mor, S. Garhwal, and A. Kumar, "A systematic review of hidden Markov models and their applications," *Archives of computational methods in engineering,* vol. 28, pp. 1429-1448, 2021.

[34] J. P. Doppler, L. C. Günther, and C. Haar, "Double-stage methodology for activity recognition in manual assembly," *Procedia CIRP,* vol. 104, pp. 423-428, 2021.

[35] M. H. Kabir, M. R. Hoque, K. Thapa, and S.-H. Yang, "Two-layer hidden Markov model for human activity recognition in home environments," *International Journal of Distributed Sensor Networks,* vol. 12, no. 1, p. 4560365, 2016.

[36] X. Cheng, B. Huang, and J. Zong, "Device-free human activity recognition based on GMM-HMM using channel state information," *IEEE Access,* vol. 9, pp. 76592-76601, 2021.

[37] M. Belgiu and L. Drăguţ, "Random forest in remote sensing: A review of applications and future directions," *ISPRS journal of photogrammetry and remote sensing,* vol. 114, pp. 24-31, 2016.

[38] A. Wang, H. Chen, C. Zheng, L. Zhao, J. Liu, and L. Wang, "Evaluation of random forest for complex human activity recognition using wearable sensors," in *2020 International Conference on Networking and Network Applications (NaNA)*, 2020: IEEE, pp. 310-315.

[39] C. Dewi and R.-C. Chen, "Human activity recognition based on evolution of features selection and random Forest," in *2019 IEEE international conference on systems, man and cybernetics (SMC)*, 2019: IEEE, pp. 2496-2501.

[40] L. Xu, W. Yang, Y. Cao, and Q. Li, "Human activity recognition based on random forests," in *2017 13th international conference on natural computation, fuzzy systems and knowledge discovery (ICNC-FSKD)*, 2017: IEEE, pp. 548-553.

[41] S. Balli, E. A. Sağbaş, and M. Peker, "Human activity recognition from smart watch sensor data using a hybrid of principal component analysis and random forest algorithm," *Measurement and Control,* vol. 52, no. 1-2, pp. 37-45, 2019.

[42] M. A. Wani, F. A. Bhat, S. Afzal, and A. I. Khan, *Advances in deep learning*. Springer, 2020.

[43] A. Mathew, P. Amudha, and S. Sivakumari, "Deep learning techniques: an overview," *Advanced Machine Learning Technologies and Applications: Proceedings of AMLTA 2020,* pp. 599-608, 2021.

[44] A. Tasdelen and B. Sen, "A hybrid CNN-LSTM model for pre-miRNA classification," *Scientific reports,* vol. 11, no. 1, pp. 1-9, 2021.

[45] W. Fang, Y. Chen, and Q. Xue, "Survey on research of RNN-based spatio-temporal sequence prediction algorithms," *Journal on Big Data,* vol. 3, no. 3, p. 97, 2021.

[46] J. Cao, F. Guo, X. Lai, Q. Zhou, and J. Dai, "A Tree-structure convolutional neural network for temporal features exaction on sensor-based multi-resident activity recognition," in *International Conference on Neural Computing for Advanced Applications*, 2020: Springer, pp. 513-525.

[47] J. Xiao and Z. Zhou, "Research progress of RNN language model," in *2020 IEEE International Conference on Artificial Intelligence and Computer Applications (ICAICA)*, 2020: IEEE, pp. 1285-1288.

[48] S. Ashry, R. Elbasiony, and W. Gomaa, "An LSTM-based descriptor for human activities recognition using IMU sensors," in *Proceedings of the 15th International Conference on Informatics in Control, Automation and Robotics, ICINCO*, 2018, vol. 1, pp. 494-501.

[49] D. Liciotti, M. Bernardini, L. Romeo, and E. Frontoni, "A sequential deep learning application for recognising human activities in smart homes," *Neurocomputing,* vol. 396, pp. 501-513, 2020.

[50] A. Natani, A. Sharma, and T. Perumal, "Sequential neural networks for multi-resident activity recognition in ambient sensing smart homes," *Applied Intelligence,* vol. 51, no. 8, pp. 6014-6028, 2021.

[51] A. Gumaei, M. M. Hassan, A. Alelaiwi, and H. Alsalman, "A hybrid deep learning model for human activity recognition using multimodal body sensing data," *IEEE Access,* vol. 7, pp. 99152-99160, 2019.

[52] D. Chen, S. Yongchareon, E. M. K. Lai, J. Yu, Q. Z. Sheng, and Y. Li, "Transformer With Bidirectional GRU for Nonintrusive, Sensor-Based Activity Recognition in a Multiresident Environment," *IEEE Internet of Things Journal,* vol. 9, no. 23, pp. 23716-23727, 2022, doi: 10.1109/jiot.2022.3190307.

[53] C. Xu, D. Chai, J. He, X. Zhang, and S. Duan, "InnoHAR: A deep neural network for complex human activity recognition," *Ieee Access,* vol. 7, pp. 9893-9902, 2019.

[54] A. Benmansour, A. Bouchachia, and M. Feham, "Multioccupant activity recognition in pervasive smart home environments," *ACM Computing Surveys (CSUR),* vol. 48, no. 3, pp. 1-36, 2015.

[55] R. Mohamed, T. Perumal, M. N. Sulaiman, and N. Mustapha, "Multi resident complex activity recognition in smart home: a literature review," *Int. J. Smart Home,* vol. 11, no. 6, pp. 21-32, 2017.

[56] R. Naccarelli, S. Casaccia, and G. M. Revel, "The Problem of Monitoring Activities of Older People in Multi-Resident Scenarios: An Innovative and Non-Invasive Measurement System Based on Wearables and PIR Sensors," *Sensors,* vol. 22, no. 9, p. 3472, 2022.

[57] L. Arrotta, C. Bettini, and G. Civitarese, "MICAR: multi-inhabitant context-aware activity recognition in home environments," *Distributed and Parallel Databases,* pp. 1-32, 2022.

[58] T. Wang and D. J. Cook, "smrt: Multi-resident tracking in smart homes with sensor vectorization," *IEEE transactions on pattern analysis and machine intelligence,* vol. 43, no. 8, pp. 2809-2821, 2020.

[59] P. Lapointe, K. Chapron, and K. Bouchard, "A new device to track and identify people in a multi-residents context," *Procedia Computer Science,* vol. 170, pp. 403-410, 2020.

[60] M. A. U. Alam, N. Roy, and A. Misra, "Tracking and behavior augmented activity recognition for multiple inhabitants," *IEEE Transactions on Mobile Computing,* vol. 20, no. 1, pp. 247-262, 2019.

[61] L. Song and Y. Wang, "Multiple target counting and tracking using binary proximity sensors: Bounds, coloring, and filter," in *Proceedings of the 15th ACM international symposium on Mobile ad hoc networking and computing*, 2014, pp. 397-406.

[62] H. Alemdar, H. Ertan, O. D. Incel, and C. Ersoy, "ARAS human activity datasets in multiple homes with multiple residents," in *2013 7th International Conference on Pervasive Computing Technologies for Healthcare and Workshops*, 2013: IEEE, pp. 232-235.

[63] M. Espinilla, E. De-La-Hoz-Franco, E. R. Bernal Monroy, P. Ariza-Colpas, and F. Mendoza-Palechor, "UJA Human Activity Recognition multi-occupancy dataset," in *Proceedings of the 54th Hawaii International Conference on System Sciences*, 2021, p. 1938.

[64] R. G. Ramos, J. D. Domingo, E. Zalama, J. Gomez-Garcia-Bermejo, and J. Lopez, "SDHAR-HOME: A Sensor Dataset for Human Activity Recognition at Home," *Sensors (Basel),* vol. 22, no. 21, Oct 23 2022, doi: 10.3390/s22218109.

[65] L. Arrotta, C. Bettini, and G. Civitarese, "The marble dataset: Multi-inhabitant activities of daily living combining wearable and environmental sensors data," in *International Conference on Mobile and Ubiquitous Systems: Computing, Networking, and Services*, 2021: Springer, pp. 451-468.

[66] D. Riboni and F. Murru, "Unsupervised recognition of multi-resident activities in smart-homes," *IEEE Access,* vol. 8, pp. 201985-201994, 2020.

[67] R. Mohamed, M. N. S. Zainudin, T. Perumal, and S. Muhammad, "Adaptive Profiling Model for Multiple Residents Activity Recognition Analysis Using Spatio-temporal Information in Smart Home," in *Proceedings of the 8th International Conference on Computational Science and Technology*, 2022: Springer, pp. 789-802.

[68] J. Cao, J. Chu, F. Guo, K. Liu, R. Xie, and H. Qin, "Ftmar: A Fusion Transformer Network for Multi-Resident Activity Recognition," *Available at SSRN 4064632*.

[69] Y. Zhan and H. Haddadi, "MoSen: Sensor Network Optimization in Multiple-Occupancy Smart Homes," in *2021 IEEE International Conference on Pervasive Computing and Communications Workshops and other Affiliated Events (PerCom Workshops)*, 2021: IEEE, pp. 384-388.

[70] D. Chen, S. Yongchareon, E. M.-K. Lai, J. Yu, and Q. Z. Sheng, "Hybrid fuzzy c-means CPD-based segmentation for improving sensor-based multiresident activity recognition," *IEEE Internet of Things Journal,* vol. 8, no. 14, pp. 11193-11207, 2021.

[71] Q. Li *et al.*, "Multi-resident type recognition based on ambient sensors activity," *Future Generation Computer Systems,* vol. 112, pp. 108-115, 2020.

[72] H. Kim and D. Lee, "ADELA: attention based deep ensemble learning for activity recognition in smart collaborative environments," in *Proceedings of the 36th Annual ACM Symposium on Applied Computing*, 2021, pp. 436-444.